\documentclass[copyright,noncommercial]{eptcs}
\pdfoutput=1
\providecommand{\volume}{5}
\providecommand{\anno}{2009}
\providecommand{\firstpage}{69}
\providecommand{\eid}{6}
\providecommand{\event}{Y. Deville and C. Solnon (Eds): Sixth International Workshop\\
 on Local Search Techniques in Constraint Satisfaction (LSCS'09).}
\usepackage{breakurl}        
\usepackage{graphicx}

\title{A Constraint-directed Local Search Approach to Nurse Rostering Problems}
\author{Fang He \& Rong Qu
\institute{Automated Scheduling, Optimization and Planning (ASAP) Group}
\institute{School of Computer Science\\
University of Nottingham, Nottingham\\
Nottingham, UK}
\email{fxh\&rxq@cs.nott.ac.uk}
}
\def\titlerunning{A Constraint-directed Local Search Approach to Nurse Rostering Problems}
\def\authorrunning{Fang He \& Rong Qu}
\begin{document}
\maketitle

\begin{abstract}
In this paper, we investigate the hybridization of constraint programming and local search techniques within a large neighbourhood search scheme for solving highly constrained nurse rostering problems. As identified by the research, a crucial part of the large neighbourhood search is the selection of the fragment (neighbourhood, i.e. the set of variables), to be relaxed and re-optimized iteratively. The success of the large neighbourhood search depends on the adequacy of this identified neighbourhood with regard to the problematic part of the solution assignment and the choice of the neighbourhood size. We investigate three strategies to choose the fragment of different sizes within the large neighbourhood search scheme. The first two strategies are tailored concerning the problem properties. The third strategy is more general, using the information of the cost from the soft constraint violations and their propagation as the indicator to choose the variables added into the fragment. The three strategies are analyzed and compared upon a benchmark nurse rostering problem. Promising results demonstrate the possibility of future work in the hybrid approach.
\end{abstract}

\section{Introduction}

Nurse rostering problems (NRPs) represent an important administration activity in real world modern hospitals. It consists of assigning a set of shifts of different types to a limited number of nurses with different working skills and working contracts, while satisfying a large set of hospital rules, working practices, legislations and personal preferences. Solving nurse rostering problems properly, while concerning the efficient allocations of the limited resources available, has a positive impact on nurses' working conditions, which are strongly related to the quality level of the healthcare.

Beside the importance of the practical aspects, solving complex NRPs which is NP-hard [1] also raises a scientific challenge to the researchers. NRPs are a special type of scheduling problem with a wide range of heterogeneous and specific constraints, thus are over-constrained and hard to solve efficiently. It has been extensively studied in Operational Research, Artificial Intelligence and local search (meta-heuristics) communities for more than 40 years [2-4]. Exact procedures, in particular Operational Research techniques such as linear programming [5], integer programming [6] and mixed-integer programming [7] have been proposed to tackle the problems. Another exact procedure, constraint programming, which originated from Artificial Intelligence research, also forms an important research direction in solving NRPs [8]. Its flexibility of modelling the complex logical constraints makes constraint programming a strong candidate to model and solve NRPs. However, due to the exponential growth of search space along with the problem size, exact procedures including constraint programming are computationally expensive for solving large scale nurse rostering problems.

On the other hand, local search approaches including tabu search [10] and simulated annealing [11], etc, are shown to be highly effective for solving large scale personnel scheduling problems. They are usually applied to improve initial complete solutions obtained from a construction phase in terms of the cost of the solutions, i.e. the value of the objective function. Roughly speaking, we can categorize local search approaches into two types: traditional local search and constraint-directed local search approaches.

Traditional local search approaches employ neighbourhood moving operators, such as 2-opt, k-opt and ejection chain, etc, while the satisfaction of constraints are checked at each move [3]. The moves considered are usually only those preserving the feasibility, i.e. all constraints have been satisfied. These methods are thus lack of flexibility when moving in the search space because the search is more likely to be stuck in the region around the initial solution. Meyer auf'm Hofe [12] highlights this limitation by using a specific example where the swap neighbourhood operator is unlikely to remove a particular violation, as it requires a simultaneous change of eight specific variables. One solution to resolve this limitation is to design more complex neighbourhood moves which amend larger parts of the current assignment. Dowsland [10] employs two types of ejection chains which consist of a sequence of on/off day swaps. Louw et al. [13] employ compound moves which are similar to the idea of chain swaps. These complex moves, although highly effective, face the potential problem of exploring an exponentially large neighbourhood. Complex moves which remain polynomial time solvable are highly desired, and usually are tailored to the specific properties of the problems in hand. 

In constraint directed local search approaches, a more general neighbourhood move can be performed by using a complete search solver concerning all constraints involved. This is the central idea of large neighbourhood search (LNS) [9,14-17]. Inspired by the concept of the impact of a variable in integer programming techniques, Refalo in [18] proposed a general impact based strategy in constraint programming to measure the importance of a variable for a reduction of the search space. Laurent Perron in [15] studied propagation guided large neighborhood search approaches. An adaptive large neighborhood search approach was proposed in [14] to solve vehicle routing problems by adaptively choosing among a number of insertion and removal heuristics (neighborhood structures) to intensify and diversify the search.

In this paper, we develop and investigate a constraint-directed local search approach within a LNS scheme to tackle the NRPs. To the best of our knowledge, this is the first attempt to use this technique to NRPs. The two main advantages of this approach are: firstly, there is no need to design complicated neighbourhood move operators to reach more solutions in the search space; secondly, by using the branch and bound search in constraint programming, more than one variable assignment can be optimized within one improvement step. 

The paper is organized as follows. In section 2 we describe the NRPs and build the constraint programming model with primitive constraints, global hard constraints and global soft constraints. In section 3 we present the constraint-directed LNS with different strategies. In section 4 we analyze computational experiment results. Conclusions and future work are given in section 5.

\section{Problem description and modeling }
\subsection{Description of the nurse rostering problem }

The NRPs we are testing are derived from real-world problems in intensive care units at hospitals. The problem consists of assigning a predefined number of shifts of different types (i.e. Early, Day, Night, Late) to a number of nurses of different working contracts in a hospital ward over a scheduling period (i.e. four weeks). Among the set of benchmark problems, the main constraints are similar but variants exist with respect to the number of nurses, number of shift types, length of the scheduling period, and different subsets of the constraints. Constraints are usually categorised into two groups: hard constraints, which must be satisfied to get feasible solutions for use in practice, and soft constraints, which are not obligatory but are desired to be satisfied as much as possible. A common hard constraint is to assign all shifts required to the limited number of nurses, i.e. demand coverage. The violations of soft constraints are usually used to evaluate the quality of solutions, for example, if a balanced workload is allocated so that human resources are used efficiently. Examples of constraints we concern in this work are listed as follows:

The hard constraints are:
\begin{itemize}
\item To a full-time nurse, 18 shifts per scheduling period have to be assigned.
\item To a part-time nurse, 10 shifts per scheduling period have to be assigned. 
\item On each day there starts at most one shift for a nurse.
\item	Each nurse receives at most 4 night shifts, of which at most 3 on consecutive days, per scheduling period.
\item	Each nurse works at most 6 days in a row.
\item	Each any 3 consecutive weekends, at least one is free of labor.
\item	After a series of 1, 2, or 3 night shifts, there is 48 hours free of labor.
\end{itemize}
The soft constraints for a nurse are:
\begin{itemize}
\item	A single night shift causes a penalty of 100. 
\item	A stand-alone shift, i.e. a single shift with day off before and after, causes a penalty of 100. 
\item	A weekend with 1 shift causes a penalty of 100. 
\item	A single day off causes a penalty of 10. 
\item	A full-time nurse receives 4 or 5 shifts, and a part-time nurse receives 2 or 3 shifts per week (from Monday till Monday). A penalty of 1, 4, 9 and 16 occurs for a deviation of 1, 2, 3 and 4 from this range, respectively. 
\item	A full-time nurse can work a series of shifts of length 4 to 6, and a part-time nurse can work a series of shifts of length 2 or 3. A penalty of 1, 4 and 9 occurs for a deviation of 1, 2 and 3, respectively. (Too short series at the end of the scheduling period are not penalized as they are carried forward to the next scheduling period).
\end{itemize}

\subsection{Constraint programming model for the nurse rostering problem}

Following our previous research [20], we formulate the NRPs in a constraint programming model within the LNS scheme. 

Denotations:
\begin{itemize}
\item $I$: set of nurses available(index $i$)
\item $J$: number of days within the scheduling period(index $j$)
\item $K$: set of shift types(includes off shift)(index $k$)
\item $D_{jk}$:coverage demand of shift type $k$ on day $j$
\item Decision variable $s_{ij}$: the shift type assigned to nurse $i$ on day $j$
\end{itemize}

Primitive constraints are used to model the problem. For example, one extra constraint to the above list is that "after a Late shift no Early shift is allowed". It can be expressed by using "if $s_{ij} = Late$, then $s_{ij}\neq Early$".

In addition to primitive constraints, to efficiently model the features of the problem, global constraint such as cardinality(gcc) has been used in the model. For example, the demand coverage constraint can be expressed as $cardinality(x, D, LB, UB)$, restricting the decision variable $x$ to take only values in shift type set $D$ of a number of times within the bounds $[LB, UB]$. It is quite straightforward to use gcc to model the constraint crisply, i.e. the constraint is either satisfied or not satisfied [12, 20]. However, the problem we are trying to tackle is over-constrained. Using only crisp constraints to model all the above stated constraints does not usually produce feasible solutions for the problems. Therefore, global soft constraints have been used to model the originally over-constrained problems as constraint optimization problems. Global soft constraints upon group of variables can be seen as a preferential constraint whose satisfaction is not required but preferred. A cost is associated to each global soft constraint in order to quantify the violation of the corresponding constraint. The objective is defined as to minimize the total sum of the cost. 

Definition: $\mu$ is a violation measure for the global constraint $c(x_1, \cdots, x_n)$ iff $\mu$ is a function $D_1\times D_2 \times \cdots \times D_n \Rightarrow R^+ $. s.t. $\forall A \in D_1 \times D_2 \times \cdots \times D_n$, where $D_1,\cdots, D_n$ is the domain of $x_1,\cdots, x_n$. $\mu(A) = 0 $ iff tuple $A$ satisfies $c(x_1,\cdots, x_n)$.

To evaluate the violation of the soft constraint, we define the violation measure $\mu$ for soft constraint cardinality as:

\begin{eqnarray}
\mu_{(soft-cardinality)}=card - UB, \quad if \quad   card > UB\\ 
\                       =LB - card, \quad 	if \quad   card < LB \nonumber
\end{eqnarray}

For example, due to the constraint that nurse $i$ should work 4-5 Day shifts in one week, assigning the nurse Day shifts of less or over the range [4, 5] will cause a penalty of 100. For a given schedule:
$l$ = Day, Day, Day, Off, Off, Off, Off, we have $Cost(l) = w\times\mu(l) = 100\times (4 - 3) = 100$, where $card(l) = 3, LB = 4, UB = 5$. 

In this paper we use the global constraint gcc and its soft version at different stages of our solution procedure. In the first stage, only hard constraints are considered and they are modeled as crisp constraints to obtain initial solutions. In the second stage, local search is applied to improve the initial solution with respect to global soft constraints. The soft gcc is used not only in the objective function, but also to decide the structure of neighborhood in the LNS approach. That is, we choose variables to be added to the fragment if these variables are linked and propagated by these soft global constraints.

\section{The constraint-directed large neighborhood search approach}

LNS is firstly proposed by Shaw in [17] to solve the vehicle routing problem. The framework of LNS is a very simple scheme in the context of constraint programming. The basic idea is to iteratively relax and then re-optimize a part of the solution assignment to find better solutions over iterations. Constraint programming is used to generate the new assignment for the relaxed part of variables and to add bound to the search to ensure that the new solution found is better than the current one.
 
Table 1 presents our 2-stage approach to solve NRPs. In the first stage, a feasible initial solution is constructed with respect to the hard constraints. In the second stage, LNS is used to iteratively improve the initial solution. The LNS is parameterized with different strategies to choose the low quality fragment (poor assignment) in the solution, which is relaxed and re-optimized to obtain improved solutions iteratively.

\begin{table}
\caption{The large neighborhood search scheme}
\begin{tabular}{l}
\hline\noalign{\smallskip}
Stage 1: construct initial solution: solve (H)  	// H: set of hard constraints\\
Stage 2: \emph {While} optimal solution not found or stopping criteria not met \emph do //LNS\\
\hspace{2 cm}Choose the low quality fragment to be relaxed \\
\hspace{2 cm}Freeze the remaining variables\\
\hspace{2 cm}Re-optimize the fragment with strategy $i$ // see sections 3.1-3.3\\
\hspace{2 cm}If found (a) First improved solution, or\\
\hspace{3.3 cm} (b) Best improved solution // search heuristics, see section 3.4\\
\hspace{3.3 cm} Update solution\\
\hspace{2 cm} End if\\
\hspace{1.5 cm}End while \\
\noalign{\smallskip}\hline
\end{tabular}
\end{table}
As observed by several researchers [14, 15], the key issue in designing efficient LNS is the selection of the fragment, i.e. set of variables to be relaxed and re-optimized. For instance, in job shop scheduling problems, the fragments usually involve the critical path of the schedules. In routing problems, cluster removing techniques have been used. To a certain extent, the success of LNS depends on the adequacy of this fragment (neighbourhood) with regard to the problematic parts of the solution and the choice of the neighbourhood size.
 
The roster (solution) we construct for NRPs has a 2-Dimensional row/column structure. Each row represents the schedule for a nurse and each column represents a day assignment in the scheduling period. The constraints involved in the model can thus be categorized as row/horizontal constraints and column/vertical constraints. In our problem, there is only one hard constraint, coverage constraint, which can be seen as a column constraint. All other constraints concerning shift patterns and preferences can be seen as row constraints. This 2-Dimensional structure of rosters determines the basic structure of the fragment selected. That is, we can select related variables by choosing which row and which column of variables to be added to the fragment.

The advantages of choosing small fragments are that complete assignment can be searched quickly. However, it is more likely that an improved solution exists in larger fragments. There is thus a trade-off between the computational time and the optimality while we set the size of the chosen fragment. Ideally the fragment should be large enough to have a reasonable chance of moving to an improved solution, and small enough so computational time is affordable. 

We investigate three strategies in this paper to choose the fragment of variables. The first two strategies are based on the observations that specific neighbourhoods with domain knowledge are usually helpful in finding good results. The first basic strategy iteratively chooses all rows (schedules of all nurses) with a fixed length over the roster, i.e. a sliding window covering fixed days/columns of the roster for all nurses. The second strategy considers the overlap of these sliding windows. In addition, a more general third strategy is investigated by using the information gained via constraint propagation to define the fragment (neighbourhood). It chooses the fragment with respect to the cost of global soft constraints, and thus is more general compared with the first two strategies.

\subsection{Fixed length sliding window as the neighborhood}

In all the constraints listed in section 2.1, the only constraint imposed on the columns is the coverage constraint cardinality. In our first strategy, we simply set the total number of nurses as the size of rows in the fragment. Based on the domain knowledge that most pattern and sequence related constraints concern one week length, i.e. full-time nurses work 4 or 5 shifts per week and part-time nurses work 2 or 3 shifts per week, etc, we set the length of rows in the fragment as 7 (days), as shown in Fig 1. This iterative selection of fragments can be seen as a sliding window with a fixed length over the roster in each iteration of the LNS loop. This is a basic and static strategy in choosing the fragment to relax and re-optimize in the LNS scheme.

\begin{figure}[htp]
  \centering
  \includegraphics[width=0.8\textwidth,height=0.3\textwidth]{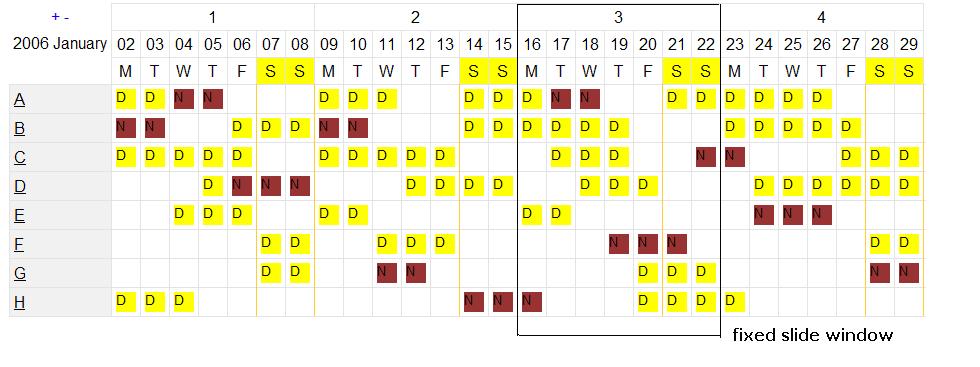}
 \caption{Fixed length sliding window as the fragment (neighborhood) in the LNS approach}
\end{figure}

\subsection{Sliding window with an overlap as the neighborhood}
Due to the row constraints concerning shift patterns and sequences, optimizing only the variables within the fixed length sliding window usually leads to violations of constraints over the variables on the boundary of these windows, as the row constraints upon the variables over the boundary interleave with the variables outside of the window. For example, as shown in Fig 2, if we relax and re-optimize the variables within the sliding window while freezing all variables outside of the window, the shift sequence of NNN for nurse H will be seen as violating the constraint of length of consecutive night shifts. In strategy 2, we consider the overlap between these sliding windows, i.e. adding the variables over the boundaries into the fragment by including different number of related variables which are involved in those constraints over the boundary.
\begin{figure}[htp]
  \centering
  \includegraphics[width=0.8\textwidth,height=0.3\textwidth]{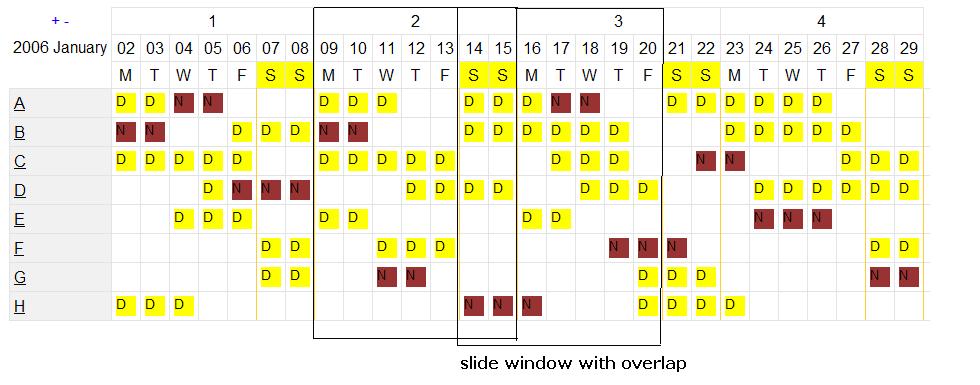}
 \caption{Sliding window with the consideration of overlap as the fragment (neighborhood) in the LNS approach}
\end{figure}

\subsection{Regions of low quality detected by cost of soft constraints}
As mentioned above, the success of LNS depends on the adequacy of neighbourhood defined with regard to the problematic part of the solution assignment and the choice of neighbourhood size. In the first two strategies, the fragment (neighborhood) is set as is set as the length of one week (columns) for all number of nurses (rows). These are efficient strategies when the problem is small, i.e. with 8 or 10 nurses. However, as the solution space increases for large problems, the computational time to re-optimize the fragment is more expensive. In strategy 3, $q$ rows of manageable size and of low quality in the roster are detected based on the cost $p(c)$ of soft constraint c. The number of rows $q$ is a thus parameter here. For example, in Fig 3, two rows (schedules for nurses C and F) with the highest cost are chosen and added to the fragment.

\begin{figure}[htp]
  \centering
  \includegraphics[width=0.8\textwidth,height=0.3\textwidth]{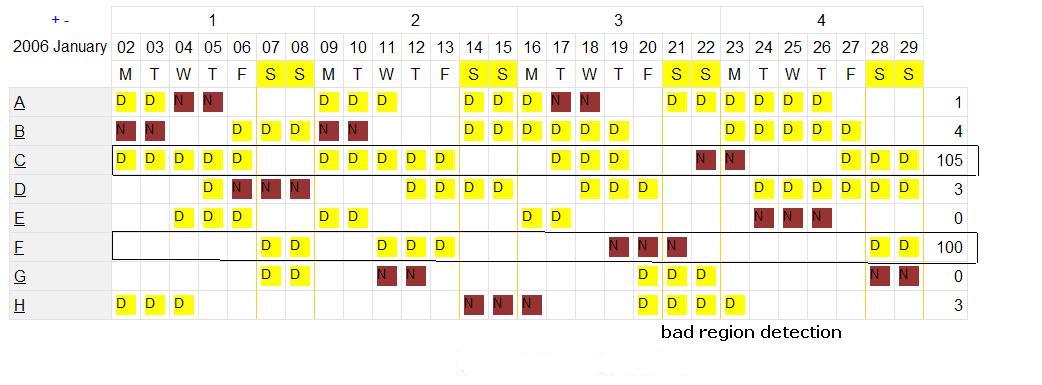}
 \caption{Region of low quality in the roster detected by constraint propagation as the fragment (neighborhood) in the LNS approach}
\end{figure}
The size of column (length of the selected rows) added to the fragment is decided by the information gained from constraint propagation. The idea is that when a variable is relaxed and re-initialized, propagation occurs. By tracing the volume of domain reduction, we can detect which variables are linked to the relaxed variables and use this information to determine the next variable to relax. The complete method using strategy 3 is described in Table 2. For each soft constraint c a variable list is maintained storing the variables linked by this constraint. Both the size of this list and the way it is updated are parameters of this algorithm. In our experiment, the variable list is consists of variables linked by the constraint involved and variables propagated by this constraint. We set the size of the list as 7 (due to the weekly structure in the roster). We also notice that the list is almost always full in the experiments, indicating that the dependencies of variables are quite tight, i.e. highly constrained. The algorithm loops until the size of fragment's search space is reasonable large.
\begin{table}
\caption{The constraint directed large neighborhood search approach using strategy 3}
\begin{tabular}{l}
\hline\noalign{\smallskip}
While Fragment size is smaller than the desired size $s$\\

\hspace{2 cm}If variable list is empty then\\
\hspace{2.5 cm}Randomly choose a linked variable\\
\hspace{2 cm}Else choose a variable in the variable list\\
\hspace{2 cm}End if \\
\hspace{2 cm}Reinitialize the chosen variable and propagate\\
\hspace{2 cm} Update the variable list: add propagated variables to the list\\
\hspace{2 cm} End if\\
End while\\
\noalign{\smallskip}\hline
\end{tabular}
\end{table}

A possible extension of the algorithm may be made to further improve the LNS search. Instead of a fixed fragment of size $s$, we can adjust the desired fragment size $s$ by adding a multiplicative correction factor to $s$, i.e. fragment of size set as $s\times\epsilon  $, where $\epsilon$  is continuously updated throughout the search process. Future investigations will be carried out to analyze the performance of the approach with different sizes of the fragment, i.e. the fixed desired size $s$ and the desired adjusted size $s\times\epsilon  $.

\subsection{Searching heuristics}
After the fragment has been chosen by either the problem specific strategies or the general strategy, the variables in the fragment are relaxed and re-optimized while all other variables are frozen to their existing assignments. The standard branch-and-bound (B\&B) search in constraint programming is used with respect to the objective function for the whole problem, i.e. sum of violations as defined in section 2.2. Within the B\&B search, we compare two variable selection heuristics to search for the optimal solution for the chosen fragment.

As shown in Table 1, when we re-optimize the chosen fragment, we can either follow the first improved rule or the best improved rule. First improved rule stops the optimization procedure as soon as the first improved solution is found; while the best improved rule solves the problem to optimal. We use the best improved rule on small size problems and the first improved rule on large size problems.

\section{Experimental results }
In this work, we evaluate our constraint-directed LNS upon a benchmark nurse rostering problem GPost, public available at http://www.cs.nott.ac.uk/$\sim$tec/NRP. The problem consists of 8 nurses who work on 3 different shift types over a scheduling period of 28 days. Other problems with different characteristics will be tested in our future work. In all experiments, 10 runs are carried out on an Intel Core 1.86GHz machine with 1.97G memory, from which average results are presented. We use the callable library in ILOG solver 6.2 as the CP solver and implement all algorithms in C++.

\subsection{Search heuristics}
We first test several variable selection heuristics to identify the efficient search heuristics in re-optimizing the chosen fragment. The variable order heuristics include: 
\begin{itemize}
\item	MinSizeInt: chooses the variable with the smallest domain
\item	MaxSizeInt: chooses the variable with the largest domain
\item	MinMinInt: chooses the variable with the least minimal bound
\item	MaxMinInt: chooses the variable with the greatest minimal bound
\item	MinMaxInt: chooses the variable with the least maximal bound
\item	MaxMaxInt: chooses the variable with the greatest maximal bound
\end{itemize}
Within the same constraint programming model for the problem, we compare in Table 3 these variable selection heuristics with respect to the number of choice points during the tree search and the number of fails encountered. Among the 6 heuristics tested, we can see that the MinSizeInt and MinMaxInt heuristics perform better than others but no statistically significant differences between themselves can be identified. We just use MinSizeInt in the follow re-optimization procedures within the LNS approach. 
\begin{table}
\caption{Evaluation of variable selection heuristics}
\begin{tabular}{llll}
\hline\noalign{\smallskip}
     & No. of choice points & No. of fails & CPU(sec)  \\
\noalign{\smallskip}\hline\noalign{\smallskip}
MinSizeInt&8966	&7995	&1.3 \\
MaxSizeInt &10706	&9723	&1.5 \\
MinMinInt&10703	&9720	&1.5\\
MaxMinInt & 11978	&10995	&1.8\\
MinMaxInt& 9290&	8319	&1.2\\
MaxMaxInt&12603	&11620	&1.8\\
\noalign{\smallskip}\hline
\end{tabular}
\end{table}
 
\subsection{Comparison of the three fragment selection strategies}
The first fixed sliding window strategy chooses the fragment of variables, represented as row*column, to be relaxed and re-optimized. The size of row is the total number of nurses while the size of column is 7 and 14 (weekly length). We also set a small size of 4 as the column size for testing purposes. The best improved rule is used on this problem. Table 4 presents the cost of solutions over iterations, from which we can see that choosing large length of sliding window obtains better result as there is more chance to obtain improved solutions within a larger neighboring size of the solution space. Meanwhile the searching time spent is acceptable for this problem.
\begin{table}
\caption{Result of fixed length sliding window as the fragment over iterations in the LNS approach}
\begin{tabular}{lllllll}
\hline\noalign{\smallskip}
 Length	&Initial solution&	Iter 1&	Iter 2&	Iter 3&	Iter 4&	CPU(sec) \\
\noalign{\smallskip}\hline\noalign{\smallskip}
4	&49&	38&	33&	32&	31&	1.56\\
7	&49	&32	&31	&24&	22&	1.78 \\
14	&49	&31	&16	&10	&8&	1.80\\
\noalign{\smallskip}\hline
\end{tabular}
\end{table}

To evaluate strategy 2 concerning a variable sliding window with overlap, different lengths of the column are compared. Due to the weekly related constraints in the problems, we set the basic length of the column as 7 (days). Based on this, we also tested different lengths of 11, 13 and 15 days. The results presented in Fig 4 demonstrate that with the largest length of sliding window = 15, better improved solutions can be obtained within less number of iterations.

We test strategy 3 based on the constraint and propagation directed fragment, and compare it with the above two strategies in Fig 5. Results demonstrate that strategies 2 and 3 perform similarly, both better than strategy 1. LNS with strategy 2 and strategy 3 obtained intervening objective values over iterations, partially because the length of columns in strategy 2 is similar to that of strategy 3, but the size of rows of strategy 2 is much larger than that of strategy 3. Thus strategy 2 searches a much larger search space than that of strategy 3. In our future work we will test these strategies with adaptively adjusted parameter s on large size problems to further investigate the effect of fragment selection based on online constraint propagation during the search.

\begin{figure}[htp]
  \centering
  \includegraphics[width=0.5\textwidth,height=0.32\textwidth]{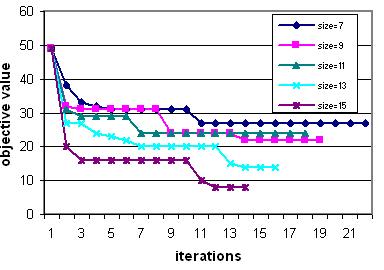}
 \caption{Results of choosing variable length sliding window with the consideration of overlap as the fragment}
\end{figure}

\begin{figure}[htp]
  \centering
  \includegraphics[width=0.5\textwidth,height=0.32\textwidth]{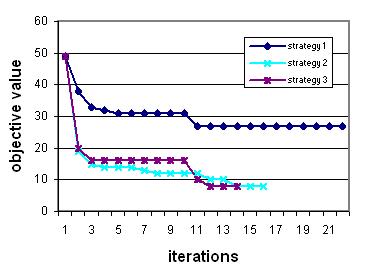}
 \caption{Comparison of the three fragment selection strategies in the LNS approach}
\end{figure}

\section{Conclusions and future work}
In this paper we investigate a constraint-directed large neighborhood search approach for solving nurse rostering problems by integrating constraint programming techniques with a local search approach. Indentified by current research in the literature, a key factor in the large neighborhood search approach is the design of neighborhoods i.e. the choice of related variables to be relaxed and re-optimized within the fragment. We focus on designing several tailored and general fragment selection strategies and compared their effect upon the performance of the large neighborhood search approach. It is shown that both the domain specific knowledge and the information of constraint propagation contribute to identifying the appropriate fragment in the large neighborhood search. Results have been analyzed to recommend future work.

In this work, we only present the preliminary results of our proposed approach applied on solving an easy problem. Further investigations are still ongoing to extend the research pursued in this paper in the following two directions:
\begin{itemize}
\item Larger benchmark nurse rostering problems will be tested to bring more insightful conclusions on our constraint directed large neighborhood search. We will also compare the approach with other existing algorithms in the literature.
\item More intelligent strategies concerning domain reduction and variable bounding, etc, will be investigated in the large neighborhood search approach. 

\end{itemize}

\bibliographystyle{eptcs} 

\end{document}